\documentclass{./styles/svproc}
\usepackage{cite}
\usepackage{amsmath,amssymb,amsfonts}
\usepackage{algorithmic}
\usepackage{graphicx}
\usepackage{textcomp}
\usepackage{xcolor}
\usepackage{url}
\usepackage{caption,subcaption}
\usepackage{multirow}
\usepackage{hyperref}

\begin{document}
\mainmatter              
\title{An Efficient Illumination Invariant Tiger Detection Framework for Wildlife Surveillance}
\titlerunning{Illumination Invariant Tiger Detection}  
%
\newcommand*\samethanks[1][\value{footnote}]{\footnotemark[#1]}
\author{Gaurav Pendharkar\thanks{Corresponding author.} \and A.Ancy Micheal\samethanks \and
Jason Misquitta \and Ranjeesh Kaippada}
\authorrunning{Gaurav et al.} 
%
\tocauthor{Gaurav Pendharkar, A.Ancy Micheal
Jason Misquitta, Ranjeesh Kaippada}
\institute{Vellore Institute of Technology,  Chennai-600127, Tamil Nadu, India,\\
\email{\{ancymicheal.a\}@vit.ac.in,\\ \{gauravsandeep.p2020, jason.misquitta2020, ranjeesh.kaippada2020\}@vitstudent.ac.in}}

\maketitle              

\begin{abstract}
Tiger conservation necessitates the strategic deployment of multifaceted initiatives encompassing the preservation of ecological habitats, anti-poaching measures, and community involvement for sustainable growth in the tiger population. With the advent of artificial intelligence, tiger surveillance can be automated using object detection. In this paper, an accurate illumination invariant framework is proposed based on EnlightenGAN and YOLOv8 for tiger detection. The fine-tuned YOLOv8 model achieves a mAP score of 61\% without illumination enhancement. The illumination enhancement improves the mAP by 0.7\%. The approaches elevate the state-of-the-art performance on the ATRW dataset by approximately 6\% to 7\%. 
\end{abstract}
\section{Introduction}
Tigers are iconic symbols of the rich biodiversity of Asia with most of their populations being primarily concentrated in countries like India, Thailand, and Indonesia\cite{i1}. These magnificent cats are apex predators and one of the vital indicators of the health of an ecosystem. However, there are multiple challenges faced in their conservation due to habitat loss, illegal trade, poaching, and extensive tourism\cite{i2}. 

In the early twentieth century, there were more than 100,000 wild tigers in Asia which has drastically reduced to fewer than 4,000 wild tigers. This precipitous decline is mainly due to a 93\% reduction in tiger habitats as a result of deforestation and agricultural expansion. Furthermore, the poaching of tigers for the illegal trade of tiger specimens and retaliatory attacks during human-wildlife conflicts are the other reasons behind the sudden decline in the population\cite{i3}. Conservationists have made commendable progress in tiger conservation but with the advent of artificial intelligence, many of the protocols can be efficiently automated by the implementation of wildlife conversation systems\cite{i4}.

Wildlife surveillance systems are a crucial tool for the conservation of endangered species, especially tigers which are characterized by elusive behaviour. In order to automate tiger detection in wildlife surveillance, accurate object detection plays a vital role. The change in lighting is one of the major challenges for efficient object detection in wildlife surveillance. Low Illumination Images(LIIs) tend to lack clarity in the illumination of the image since they are sampled and quantized in low-light environmental conditions. Traditional non-deep learning methods do not adapt based on the pixel distribution of the image and result in extreme improvement in illumination. Many of the traditional deep-learning approaches require pair-wised supervision to train the models based on the Retinex theory\cite{i9}. However, there are a few approaches implemented by General Adversarial Networks(GANs) that improve the illumination by learning some unsupervised features. Therefore, in this paper, the EnlightenGAN\cite{i10} is used to provide illumination enhancement to the images followed by object detection with the YOLOv8 model\cite{i11}.

\section{Related Work}
\textbf{Tiger Detection:} A fast yet efficient approach for real-time tiger detection was proposed by Kupyn and Pranchuk.\cite{l1}. TigerNet model was developed based on the Feature Pyramid Network (FPN) and uses Depthwise Separable Convolutions (DSC) with a lightweight backbone of FD-MobileNet\cite{m1}. The lightweight backbone of FD-MobileNet improves the detection speed which is commendable but also reduces the accuracy slightly. Tan et. al constructed a dataset of video clips taken by infrared cameras installed in the Northeast Tiger and Leopard National Park covering 17 species\cite{l2}. The main aim of the paper was to compare the performance of three mainstream object detection models, particularly FCOS Resnet101, YOLOv5, and Cascade R-CNN HRNet32. YOLOv5 showed the most consistent results among the three models. It obtained  88.8\%, 89.6\%, and 89.5\% accuracy at various thresholds. The overall high accuracy of detection is commendable but the model suffers from data imbalance and therefore has a significant variance in the species-wise performance. However, the models relatively performed well for Amur Tigers compared to other animals. 

Liu and Qu introduced the AF-TigerNet, a lightweight neural network designed for the real-time detection of Amur tigers\cite{l3}.  The approach involves improving feature extraction by incorporating an updated CSPNet\cite{p1} and cross-stage partial (CSP)-path aggregation network (PAN) architectures. Dertien et al proposed a novel approach, TrailGuard, which was developed to detect tigers and poachers and alert the respective government authorities\cite{l4}. This innovative system boasts an impressive response time of just 30 seconds from the moment the camera is triggered to the notification appearing on a smartphone app.

\textbf{Low Illumination Image Enhancement:} Al Sobbahi and Tekli reviewed different approaches to enhance the quality of LIIs and their effect on object detection\cite{l5}. The traditional approaches included Gamma Correction, Histogram Equalization, models based on Retinex theory, and frequency-based methods. They also discuss deep-learning approaches consisting of Encoder-Decoders, Retinex theory-based models, General Adversarial Networks (GANs)\cite{p2}, and Zero Reference models. Among all the approaches, Histogram Equalization, EnlightenGAN, and GladNet\cite{m2} were the ones independent of pairwise supervision. The relative performance of the models, their examples, and the explainable categorization were commendable but the open-source availability of the models was not discussed. Choudhury et al. highlighted the increase in the demand for surveillance in order to protect endangered species\cite{l6}. Furthermore, they also emphasize that the degraded quality of images with respect to bad contrast, high noise, reflectance, and lousy illumination severely affect object detection. They utilize the EnlightenGAN as a step of preprocessing before object detection using YOLOv3\cite{m3} for an automatic detection system. The GAN is evaluated using BRISQUE and NIQE with mAP, GIoU loss, F-measure, and objectness loss as metrics for the rhino detector.

Wang et al. experimented with different techniques to enhance detection in low-light scenarios\cite{l7}. Their exploration of different image enhancement algorithms reveals that EnlightenGAN and Zero\_DCE++\cite{m4} worked best in conjunction with the YOLOv5 model giving a precision of 0.747. The work improves the performance of the image enhancement techniques with traditional versions of YOLO models which is appreciable.

\section{Methodology}
\begin{figure}[htbp]
\centerline{\includegraphics[width=0.6\textwidth]{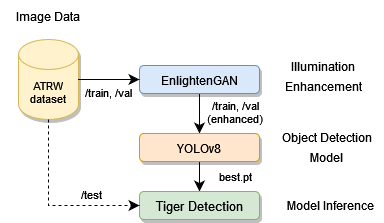}}
\caption{Architecture Diagram}
\label{arch}
\end{figure}

Illumination Variation is a tedious task during object detection. In wildlife surveillance, change in lighting conditions results in less efficient animal detection. This paper focuses on effective tiger detection by addressing illumination variation. Image restoration and enhancement methods have been evolving appreciably over the past years. However, most of the state-of-the-art deep learning approaches require pairwise training data which in most cases infeasible for real-world problems. EnlightenGAN is one such approach that uses unsupervised learning-based general adversarial networks (GANs) to devise a mapping between low-light and normal-light images. An attention-guided U-Net is used for the image generation and two discriminators namely Global and Local Discriminators. Hence, in this paper, the EnlightenGAN is applied to handle illumination variation followed by object detection with YOLOv8. The framework of the proposed architecture is shown in Figure \ref{arch}.

\subsection{EnlightenGAN}
\begin{figure}[htbp]
\centerline{\includegraphics[width=\linewidth]{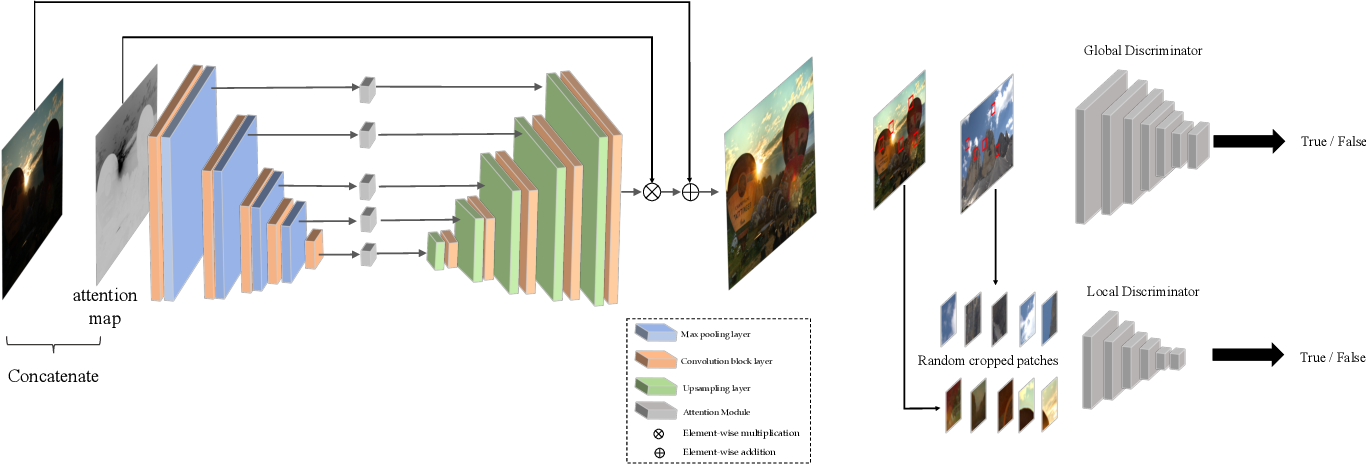}}
\caption{EnlightenGAN Architecture}
\label{gan-arch}
\end{figure}

EnlightenGAN is an unsupervised learning based GAN that uses an attention-guided unit as the generator and 2 discriminators namely the global and local discriminators. Figure \ref{gan-arch} depicts the architecture of EnlightenGAN. The model takes a low-light image as input(A) and computes its normalized illumination component(I). It computes the self-regularized map by an element-wise subtraction, 1-I. The input image and the self-regularized map are concatenated and given as input to the generator to obtain the reconstructed image($A'$). An element-wise multiplication and element-wise addition are carried out on $A'$ as shown in Eqn. \ref{gan-img-construction}. 

\begin{equation}\label{gan-img-construction}
    B = (A' \otimes I) \oplus A
\end{equation}

\begin{figure}
\centering
\begin{subfigure}[b]{2.38in}
\includegraphics[width=\linewidth]{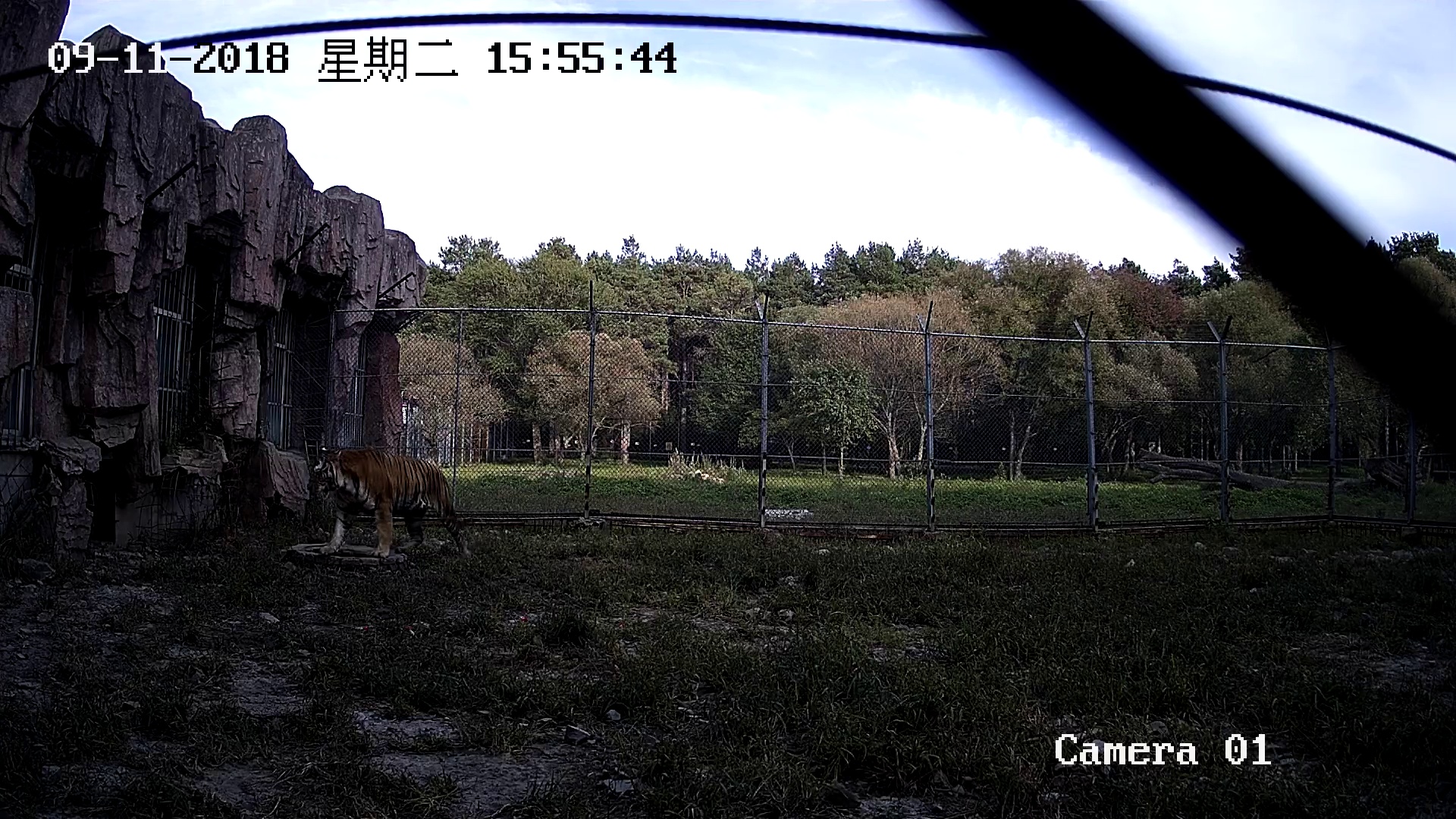}
\includegraphics[width=\linewidth]{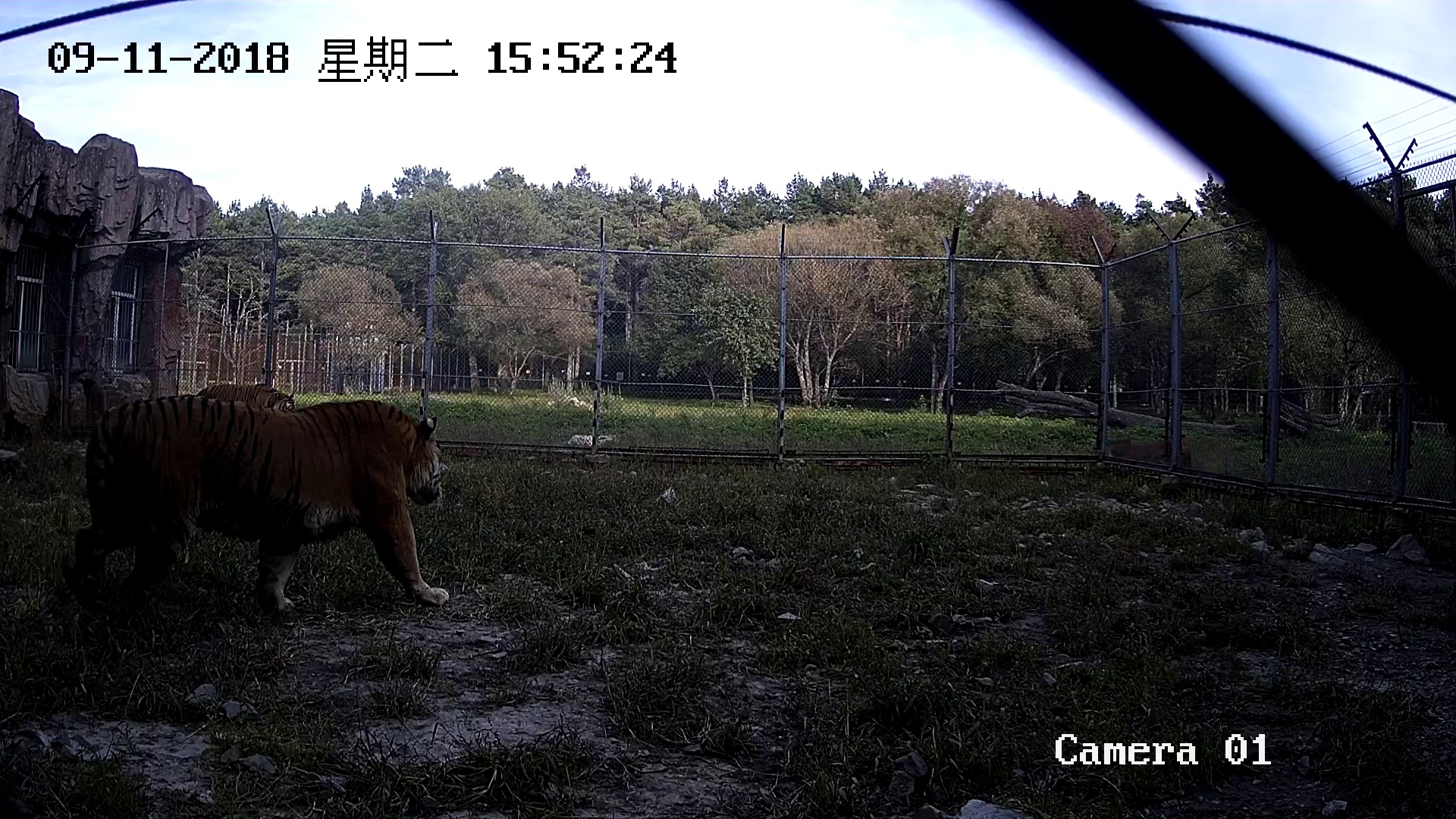}
\includegraphics[width=\linewidth]{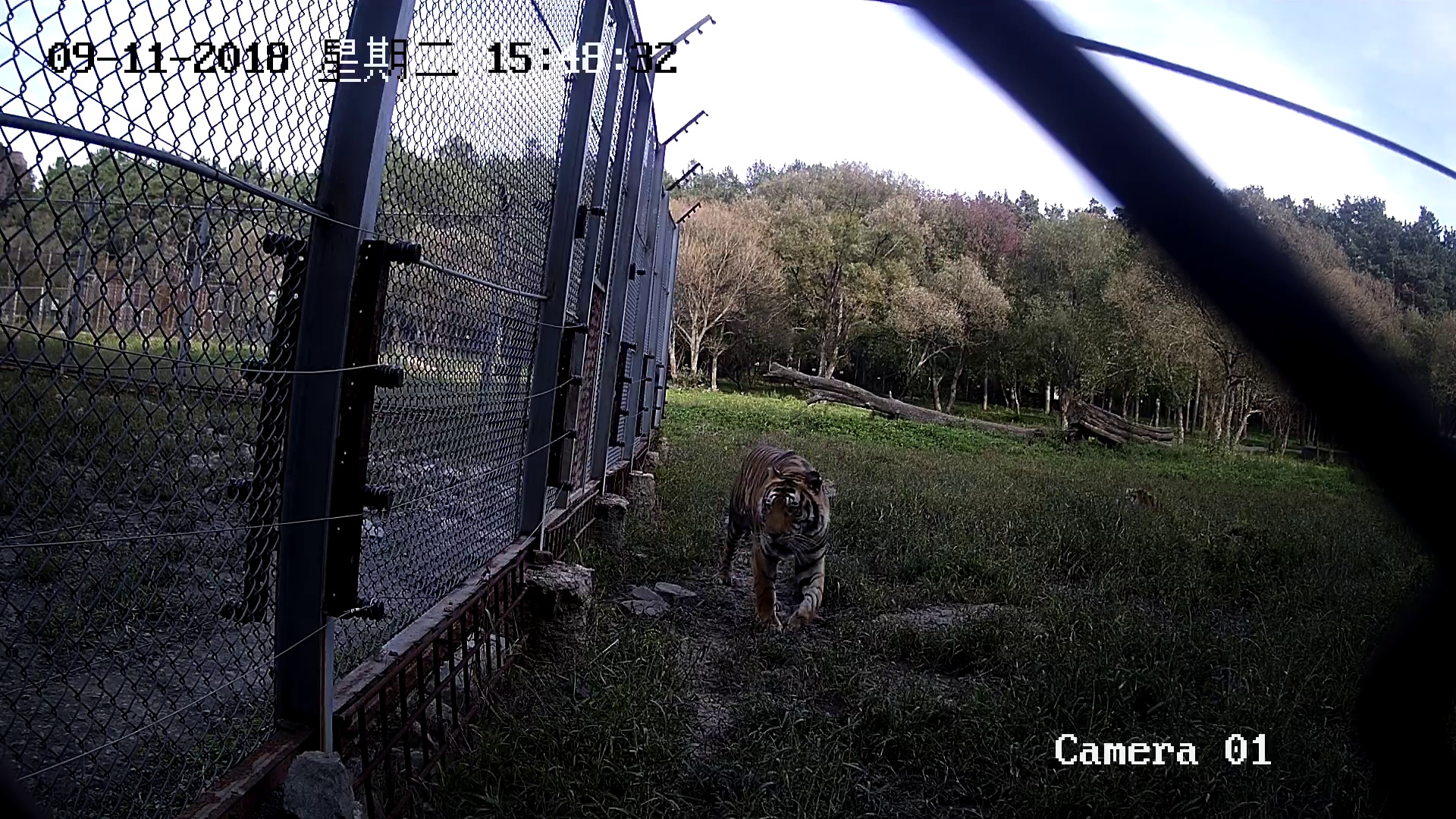}
\includegraphics[width=\linewidth]{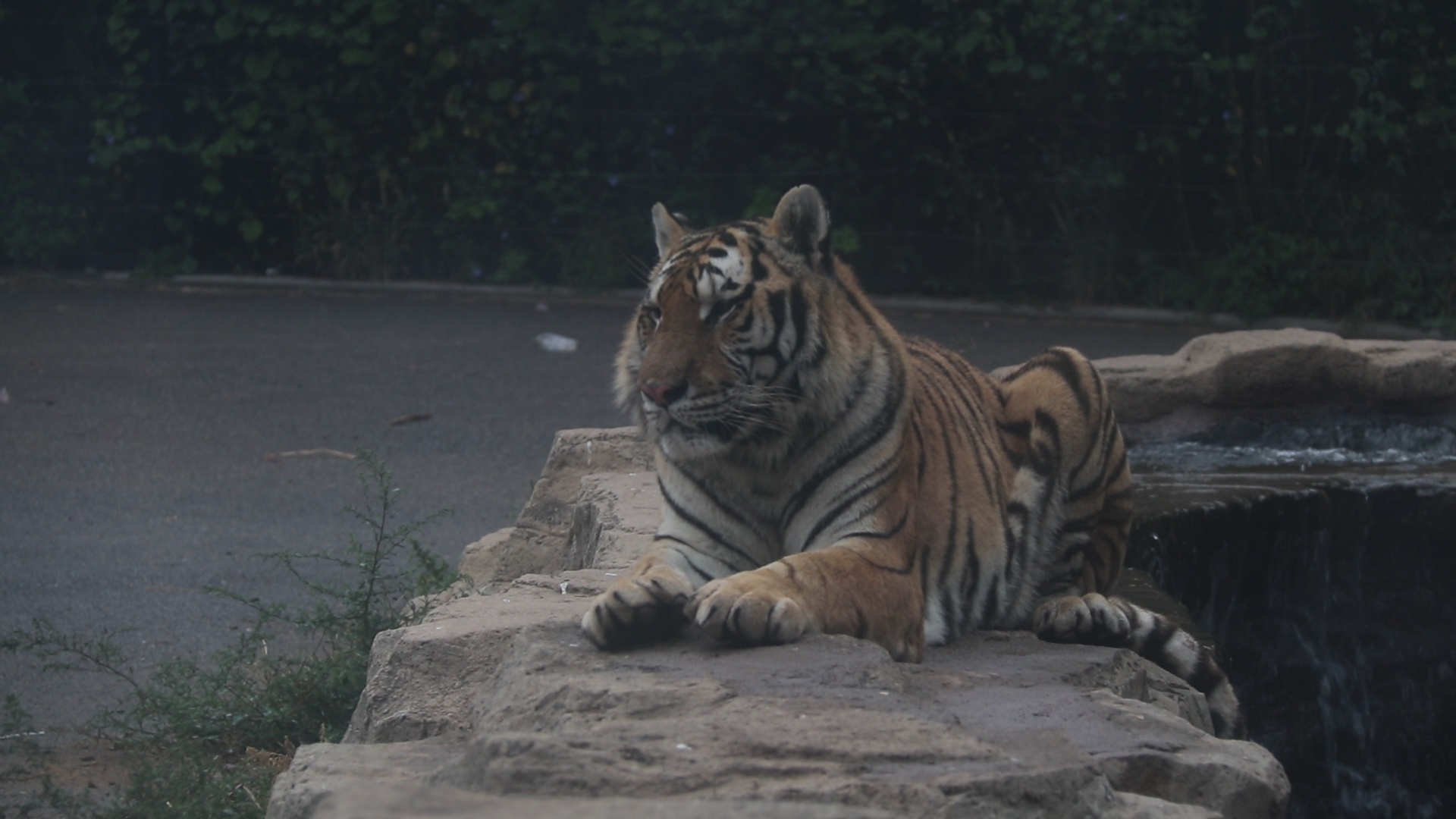}
    \caption{Original Image}
    \end{subfigure}\hfil
\begin{subfigure}[b]{2.38in}
\includegraphics[width=\linewidth]{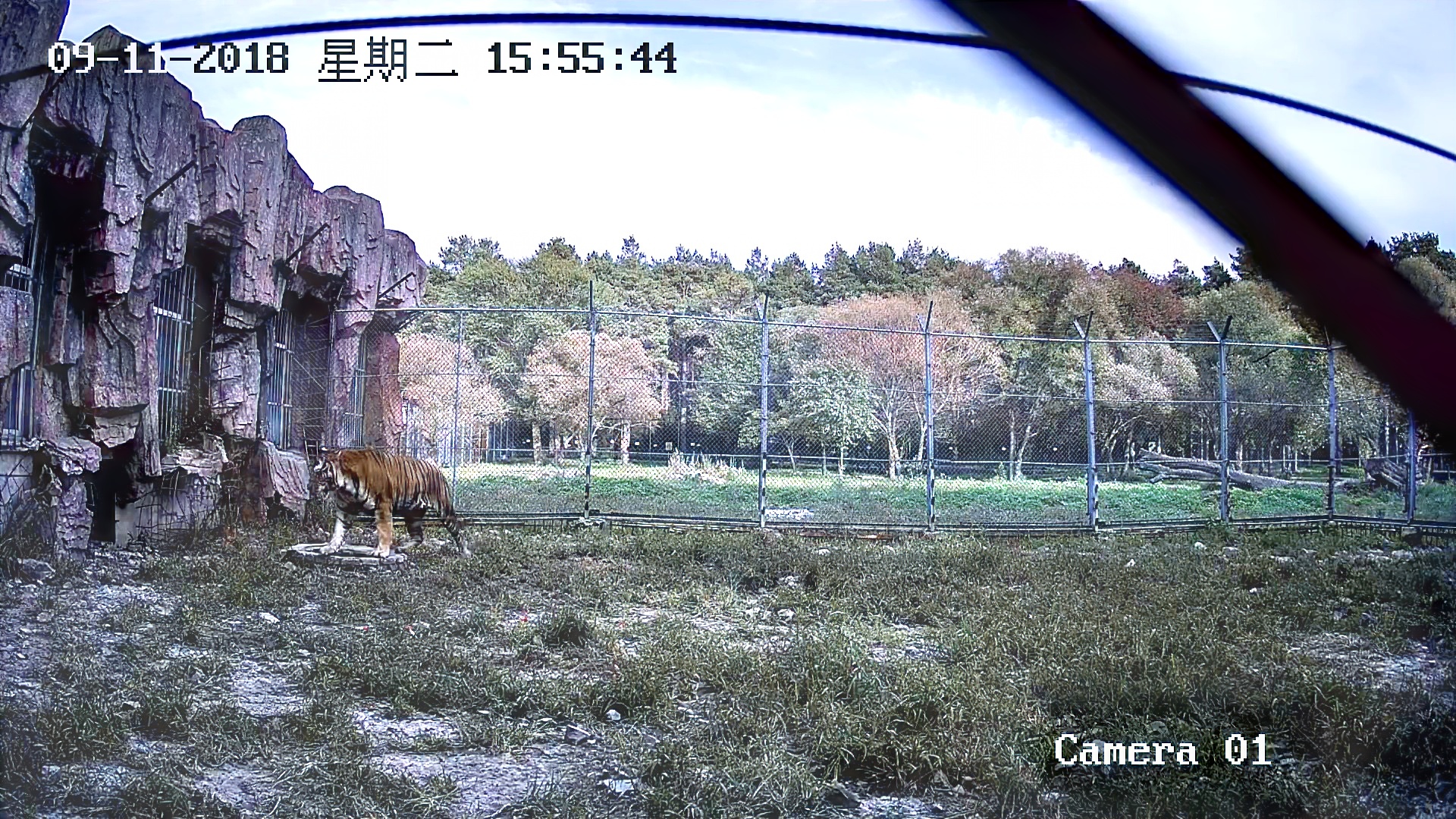}
\includegraphics[width=\linewidth]{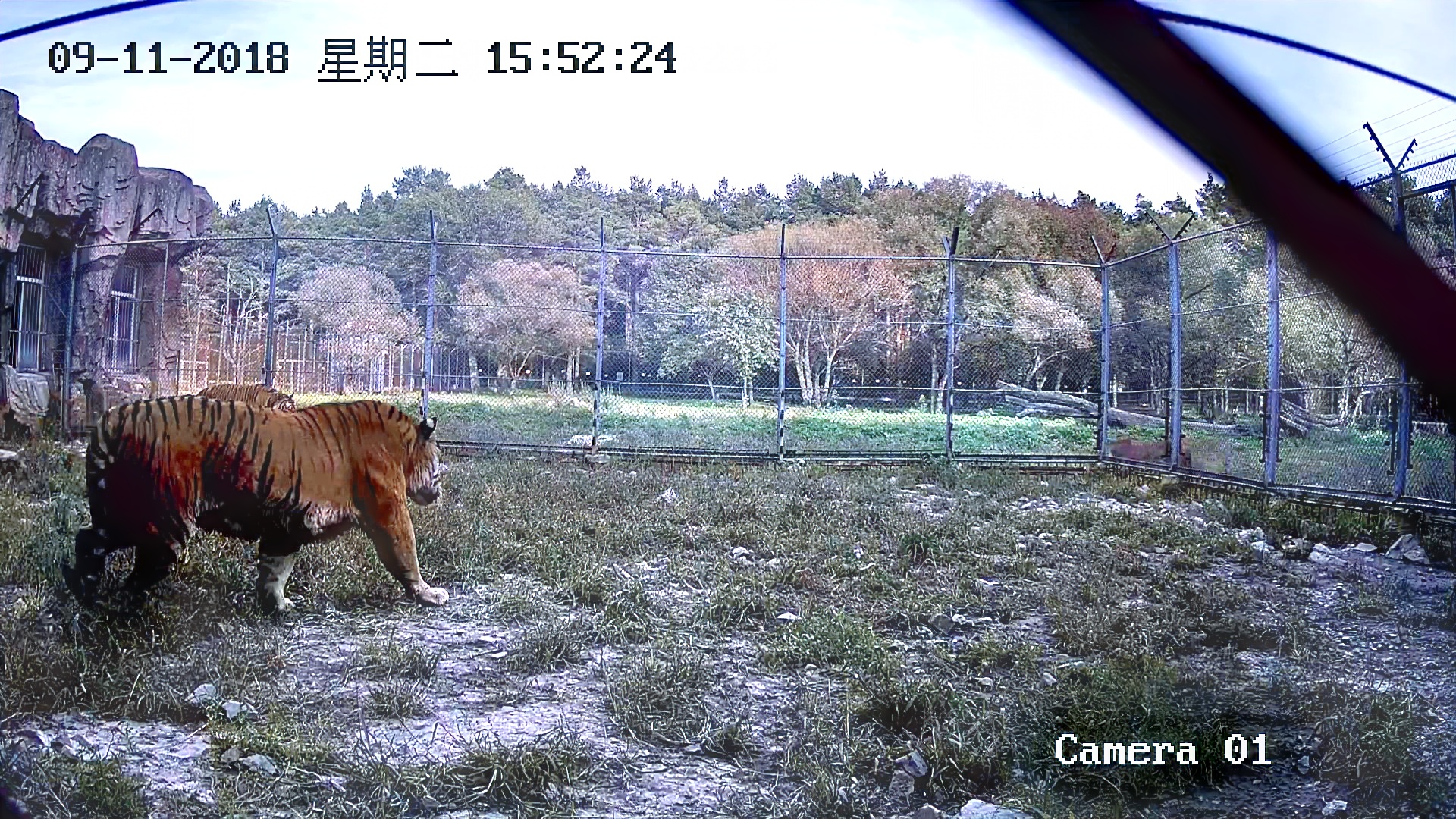}
\includegraphics[width=\linewidth]{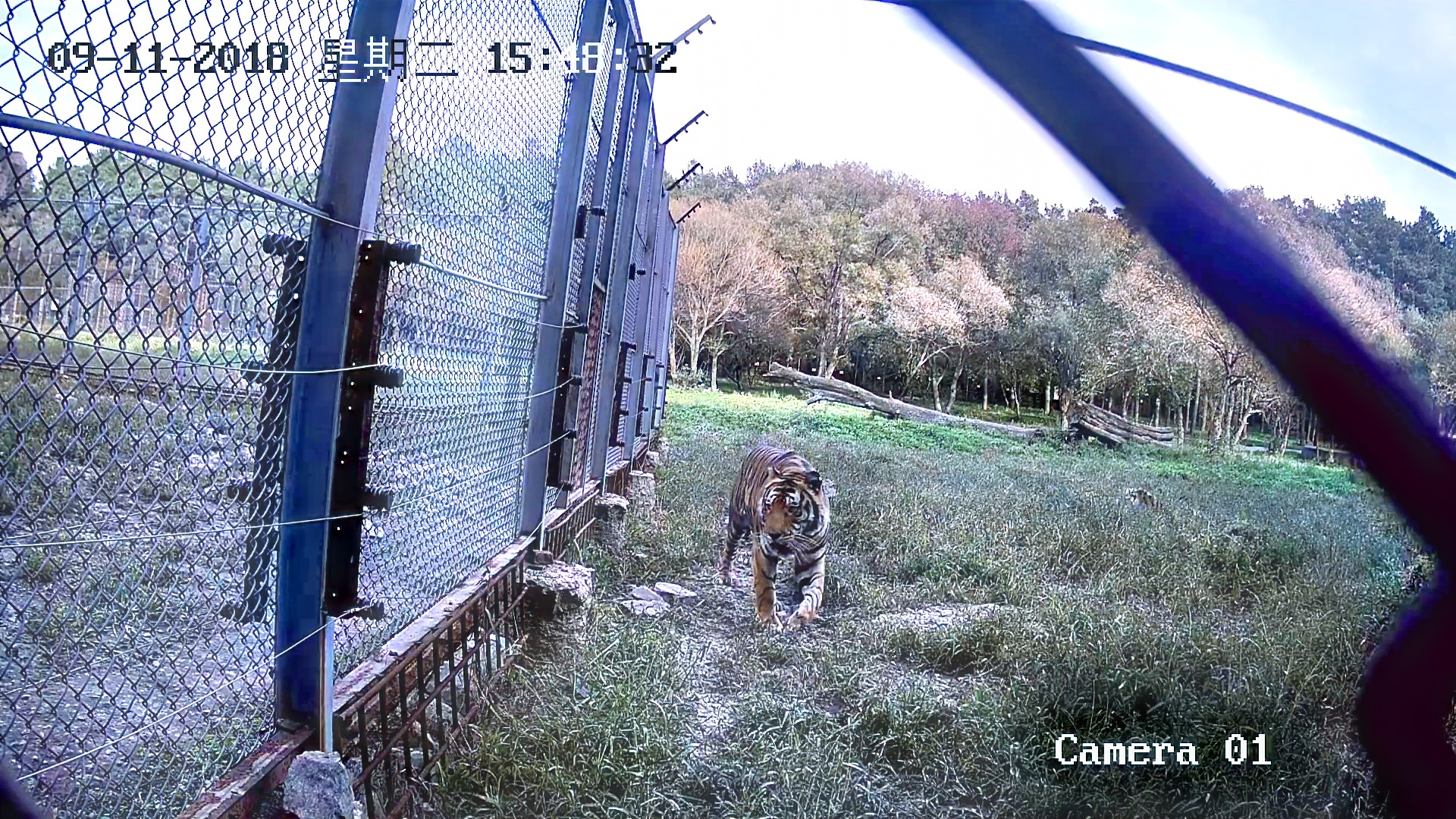}
\includegraphics[width=\linewidth]{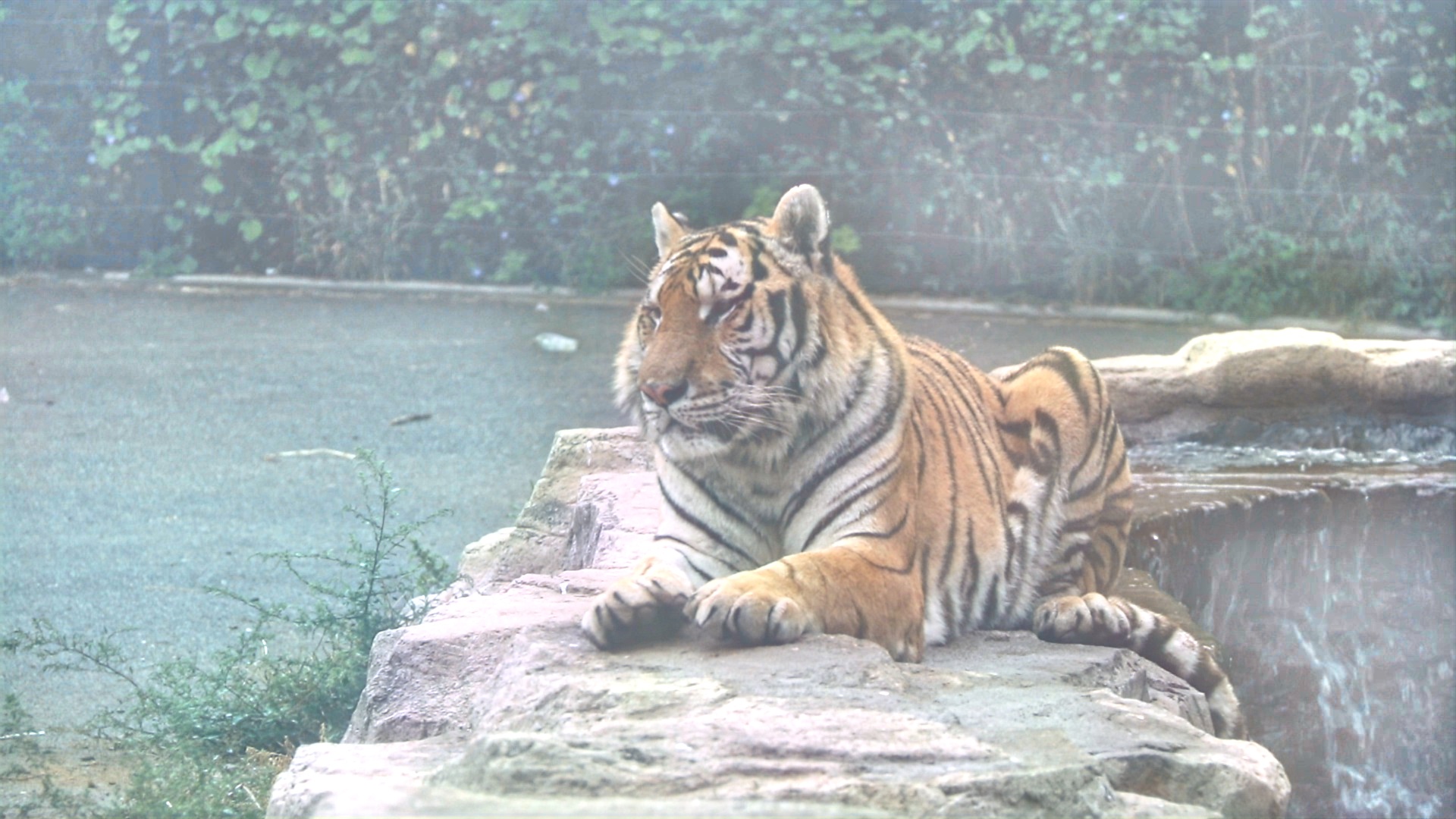}
    \caption{Illuminated Image}
    \end{subfigure}\hfil
\caption{Samples of Illuminated Image by EnlightedGAN}
\label{sample-images}
\end{figure}

The final reconstructed image(B) is given as input to the global discriminator and some randomly cropped patches are given as input to the local discriminator. Finally, both the discriminators return a true or false. The loss function used to train the network is shown in Eqn. \ref{gan-loss-func} in which $\mathcal{L}^{Global}_{SFP}$ and $\mathcal{L}^{Local}_{SFP}$ denotes the self-feature preserving loss for the global and local discriminator respectively. The loss for global and local discriminators is denoted by $\mathcal{L}^{Global}_G$ and $\mathcal{L}^{Local}_G$. The sample images illuminated by EnlightenGAN are shown in Figure \ref{sample-images}.

\begin{equation}\label{gan-loss-func}
Loss = \mathcal{L}^{Global}_{SFP} + \mathcal{L}^{Local}_{SFP} + \mathcal{L}^{Global}_G + \mathcal{L}^{Local}_G
\end{equation}

\subsection{YOLOv8}
``YOLO" an acronym for ``You Only Look Once" is a highly effective algorithm for real-time object detection, renowned for its exceptional balance between accuracy and speed. The latest iteration of the YOLO series, YOLOv8, has been created by Ultralytics. It improves the performance of the previous versions of YOLO by the following improvisations:
\begin{itemize}
    \item introduction of anchor-free detections by which the model can directly estimate the center of an object instead of relying on an offset from a known anchor box.
    \item alterations in the convolutions in the backbone network.
    \item closing mosiac augmentation before training is completed.
\end{itemize}

\begin{figure*}[htbp]
\centerline{\includegraphics[width=1.3\linewidth]{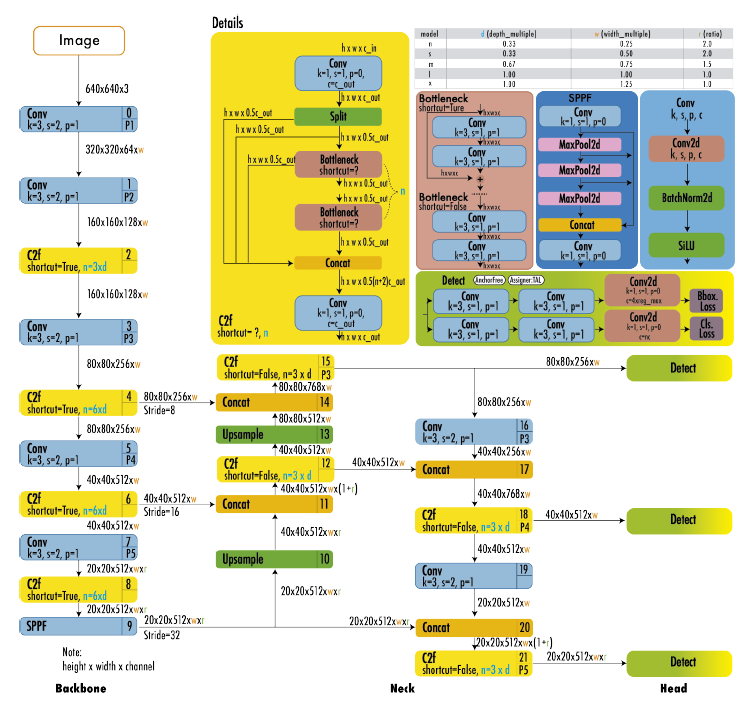}}
\caption{YOLOv8 Architecture}
\label{yolov8-arch}
\end{figure*}

\begin{figure*}[htbp]
\centerline{\includegraphics[width=1.2\linewidth]{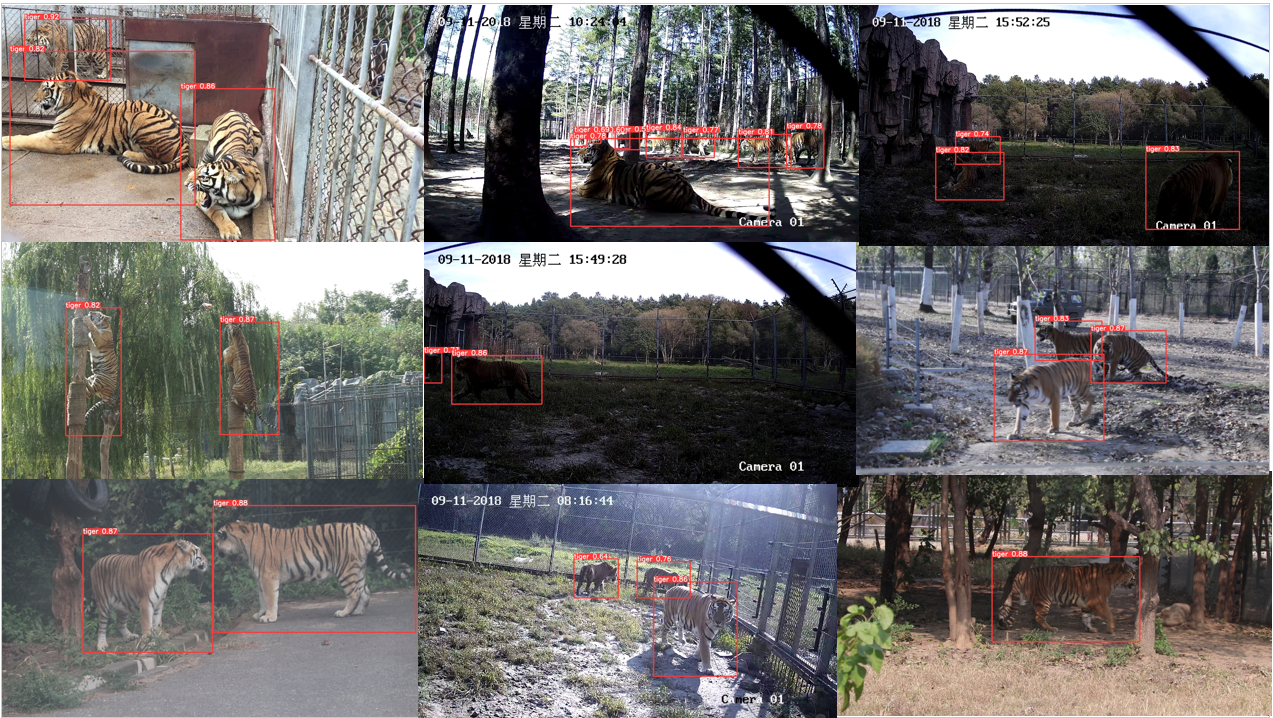}}
\caption{Sample tiger detection results by YOLOv8}
\label{yolov8-results}
\end{figure*}

Figure \ref{yolov8-arch} shows the general architecture of the YOLOv8 model which has three components namely the backbone, neck, and head. Relevant features are extracted from the input image by the backbone network. The neck connects the backbone network and the head network. It also reduces the dimensions of the feature map and improves the resolution of the features. Finally, the head network comprises three detection networks to detect small, medium, and large objects. The sample results for the YOLOv8 model are portrayed in Figure \ref{yolov8-results}.

\section{Results and Discussion}
\begin{figure*}[htbp]
\centerline{\includegraphics[width=0.8\linewidth]{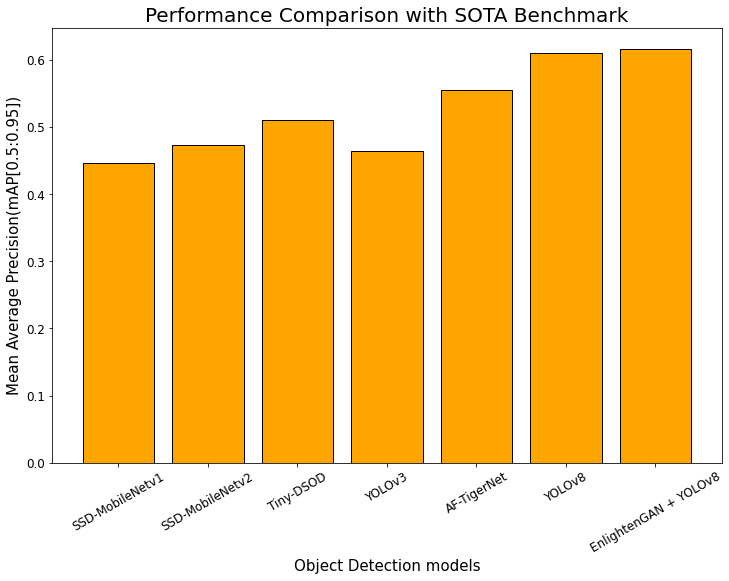}}
\caption{Pictorial representation of the comparison with SOTA performance}
\label{sota-barplot}
\end{figure*}

The experiment is performed with the ATRW dataset. ATRW dataset deals with three computer vision tasks namely tiger detection, pose estimation, and tiger re-identification. The detection dataset comprises 9496 bounding boxes across 4434 images\cite{i7}. The efficiency of the framework is evaluated using Mean Average Precision(mAP). For an object detection task, each prediction has confidence and the dimensions of the bounding box($B_{p}$). A detection is said to be correct if it satisfies the IoU threshold($t$) as shown in the Eqn. \ref{iou-threshold}. IoU stands for the Intersection over Union which is the ratio of the area of intersection to the area of union as depicted in the Eqn. \ref{iou}. Finally, among the correct predictions, the mAP is computed as per the formula given in Eqn. \ref{map}\cite{metrics}. The average Precision denoted by AP is the area under the Precision-Recall Curve. The state-of-the -art performance for the proposed methodology is shown in Table \ref{sota-comp} and is pictorially depicted in Figure \ref{sota-barplot}.
\begin{equation}\label{iou}
    IoU(B_{p}, B_{gt}) = \frac{area(B_{p} \cap B_{gt})}{area(B_{p} \cup B_{gt})}
\end{equation}

\begin{equation}
 T(B_{p}, B_{gt})=  \left\{
\begin{array}{ll}
      correct & IoU(B_{p}, B_{gt}) > t \\
      incorrect & Otherwise \\
\end{array} 
\right. 
\label{iou-threshold}
\end{equation}

\begin{equation}\label{map}
    mAP = \frac{1}{N}\sum_{i = 1}^{N} AP_{i}
\end{equation}

\begin{table}
    \caption{Comparison with SOTA Performance\cite{i7}\cite{l3}}
    \centering
    \begin{tabular}{|c|c|c|}
    \hline 
        \textbf{Model} & \textbf{mAP[0.5:0.95]} \\
    \hline
        SSD-MobileNetv1 & 0.446 \\
        SSD-MobileNetv2 & 0.473 \\
        Tiny-DSOD & 0.511 \\
        YOLOv3 & 0.464 \\
        AF-TigerNet & 0.555 \\
        \textbf{YOLOv8} & \textbf{0.610} \\
        \textbf{EnlightenGAN + YOLOv8} & \textbf{0.617} \\
        \hline 
    \end{tabular}
    \label{sota-comp}
\end{table}

The SOTA performance is 0.464 for YOLOv3. AF-TigerNet performs better than YOLOv3  with an mAP of 0.555. Further, the experiment was performed with YOLOv8 which achieved an mAP of 0.610. From Table \ref{sota-comp}, it is evident that there is a significant increase in mAP of 7\% on the application of EnlightenGAN and YOLOv8. Overall, by addressing illumination variation, the proposed framework achieves an mAP of 0.617, outperforming the other comparative models.

\section{Conclusion}
Recently, wildlife surveillance has been automated with various computer vision methodologies. Illumination variation is a tedious task in wildlife surveillance. It also affects efficient object detection during surveillance. This paper focuses on a novel framework that addresses Illumination variation and provides efficient tiger detection using EnlightenGAN and YOLOv8. The attention-guided unit, global and local discriminator in the EnlightenGAN provides illumination varied images which are further fed into the YOLOv8 object detector. The experiment is performed with the ATRW tiger dataset. The proposed model outperforms the SOTA with an mAP of 0.617. In the future, the tiger detection model can be expanded for multi-class wildlife surveillance.

\end{document}